\documentclass[sigconf]{acmart}

\usepackage{booktabs}
\usepackage{tabularx}
\usepackage{array}
\usepackage{multirow}
\usepackage{enumitem}
\usepackage{microtype}
\usepackage{tikz}
\usepackage[frozencache,cachedir=.]{minted}
\usepackage{bookmark}
\usetikzlibrary{positioning,arrows.meta,shapes.geometric,fit,backgrounds}
\AtBeginDocument{%
  }

\setcopyright{acmlicensed}
\acmYear{2026}\copyrightyear{2026}
\acmConference[CAIS '26]{Proceedings of the 1st ACM Conference on Agentic and AI Systems}{May 26, 2026}{San Jose, CA, USA}
\acmBooktitle{Proceedings of the 1st ACM Conference on Agentic and AI Systems (CAIS '26), May 26, 2026, San Jose, CA, USA}
\acmDOI{10.1145/3786335.3813206}
\acmISBN{979-8-4007-2415-2/26/05}

\begin{document}

\title{Agent Lifecycle Toolkit (ALTK): Reusable Middleware Components for Robust AI Agents}

\author{Zidane Wright}
\affiliation{%
    \institution{IBM Research}
    \city{New York}
    \state{NY}
    \country{USA}
}

\author{Jason Tsay}
\affiliation{%
    \institution{IBM Research}
    \city{New York}
    \state{NY}
    \country{USA}
}
\email{jason.tsay@ibm.com}
\authornote{Corresponding author.}

\author{Anupama Murthi}
\affiliation{%
    \institution{IBM Research}
    \city{New York}
    \state{NY}
    \country{USA}
}

\author{Osher Elhadad}
\affiliation{%
    \institution{IBM Research}
    \city{Haifa}
    \country{Israel}
}

\author{Diego Del Rio}
\affiliation{%
    \institution{IBM}
    \country{Argentina}
}

\author{Saurabh Goyal}
\affiliation{%
    \institution{IBM Research}
    \city{New York}
    \state{NY}
    \country{USA}
}

\author{Kiran Kate}
\affiliation{%
    \institution{IBM Research}
    \city{New York}
    \state{NY}
    \country{USA}
}

\author{Jim Laredo}
\affiliation{%
    \institution{IBM Research}
    \city{New York}
    \state{NY}
    \country{USA}
}

\author{Koren Lazar}
\affiliation{%
    \institution{IBM Research}
    \city{Haifa}
    \country{Israel}
}

\author{Vinod Muthusamy}
\affiliation{%
    \institution{IBM Research}
    \city{New York}
    \state{NY}
    \country{USA}
}

\author{Yara Rizk}
\affiliation{%
    \institution{IBM Research}
    \city{New York}
    \state{NY}
    \country{USA}
}


\renewcommand{\shortauthors}{Wright et al.}

\begin{abstract}
As AI agents move from demos into enterprise deployments, their failure modes become consequential: a misinterpreted tool argument can corrupt production data, a silent reasoning error can go undetected until damage is done, and outputs that violate organizational policy can create legal or compliance risk. Yet, most agent frameworks leave builders to handle these failure modes ad hoc, resulting in brittle, one-off safeguards that are hard to reuse or maintain. We present the Agent Lifecycle Toolkit (ALTK), an open-source collection of modular middleware components that systematically address these gaps across the full agent lifecycle. 

Across the agent lifecycle, we identify opportunities to intervene and improve, namely, post‑user-request, pre‑LLM prompt conditioning, post‑LLM output processing, pre‑tool validation,  post‑tool result checking, and pre‑response assembly. ALTK provides modular middleware that detects, repairs, and mitigates common failure modes. It offers consistent interfaces that fit naturally into existing pipelines. It is compatible with low-code and no-code tools such as the ContextForge MCP Gateway and Langflow. Finally, it significantly reduces the effort of building reliable, production‑grade agents.

\end{abstract}



\received{13 March 2026}


\maketitle

\section{Introduction}
The agentic paradigm has accelerated rapidly as developers build increasingly capable LLM‑powered agents that can reason, call tools, and produce structured outputs. Yet, these systems remain fundamentally brittle: as complexity grows, so do issues like hallucinated tool calls, silent failures, inconsistent outputs, and reasoning errors that break workflows. To address these challenges, we introduce ALTK, an open‑source, framework-agnostic package that improves agent reliability, predictability, and production readiness. Agent Lifecycle Toolkit (ALTK) can integrate into any agent pipeline and add deterministic safeguards and recovery mechanisms that elevate agents from “cool demos” to dependable, enterprise-grade systems.

Early agents often rely on a simple loop of repeated LLM tool calls, useful for prototypes but insufficient for enterprise reliability. Production agents need additional logic to ensure robustness, especially in domains like sales where a single misinterpreted field can trigger incorrect APIs and distort downstream forecasts. Agent orchestration frameworks such as 
LangChain \cite{chase2022langchain}, LangGraph \cite{langchain2024langgraph}, CrewAI~\cite{moura2023crewai} offer building blocks such as tools, memory, and popular agent architectures. However, they expect the developers to write custom code handling tool call errors or checking for policy conformance.

ALTK is a modular toolkit that comes with pre-built, hardened drop-in components to strengthen reasoning, tool execution, and output validation in agents. Rather than enforcing a particular agent framework (such as LangChain, LangGraph, or AutoGPT), its framework‑agnostic design allows teams to introduce targeted reliability improvements without re-architecting their agents.

ALTK currently includes 10 components, each addressing a distinct failure mode in the agent lifecycle, as summarized in Figure \ref{fig:lifecycle}. For example, the lifecycle stage of "Pre-Tool" indicates a step in the agent's execution when the LLM has generated a tool call but the tool call is yet to be executed. A "Pre-Tool" ALTK component such as SPARC takes the generated tool call, tool specifications, and agent context as input and performs different checks on the tool call to check for correctness. SPARC flags an invalid tool call with reasoning and suggestions for correcting it so the agent skips executing the wrong tool call and asks the LLM to take this feedback and generate a correct call.

What distinguishes ALTK is its flexibility through precise, surgical improvements rather than all-in-one orchestration. This allows for multiple paths of integration into existing agentic systems, including existing low-code and no-code systems such as LangFlow and the ContextForge MCP Gateway. All ALTK components have been rigorously evaluated on public benchmarks and show clear gains over baseline settings.

\section{Toolkit Approach}
\begin{figure*}[t]
\centering
    \includegraphics[width=0.9\linewidth]{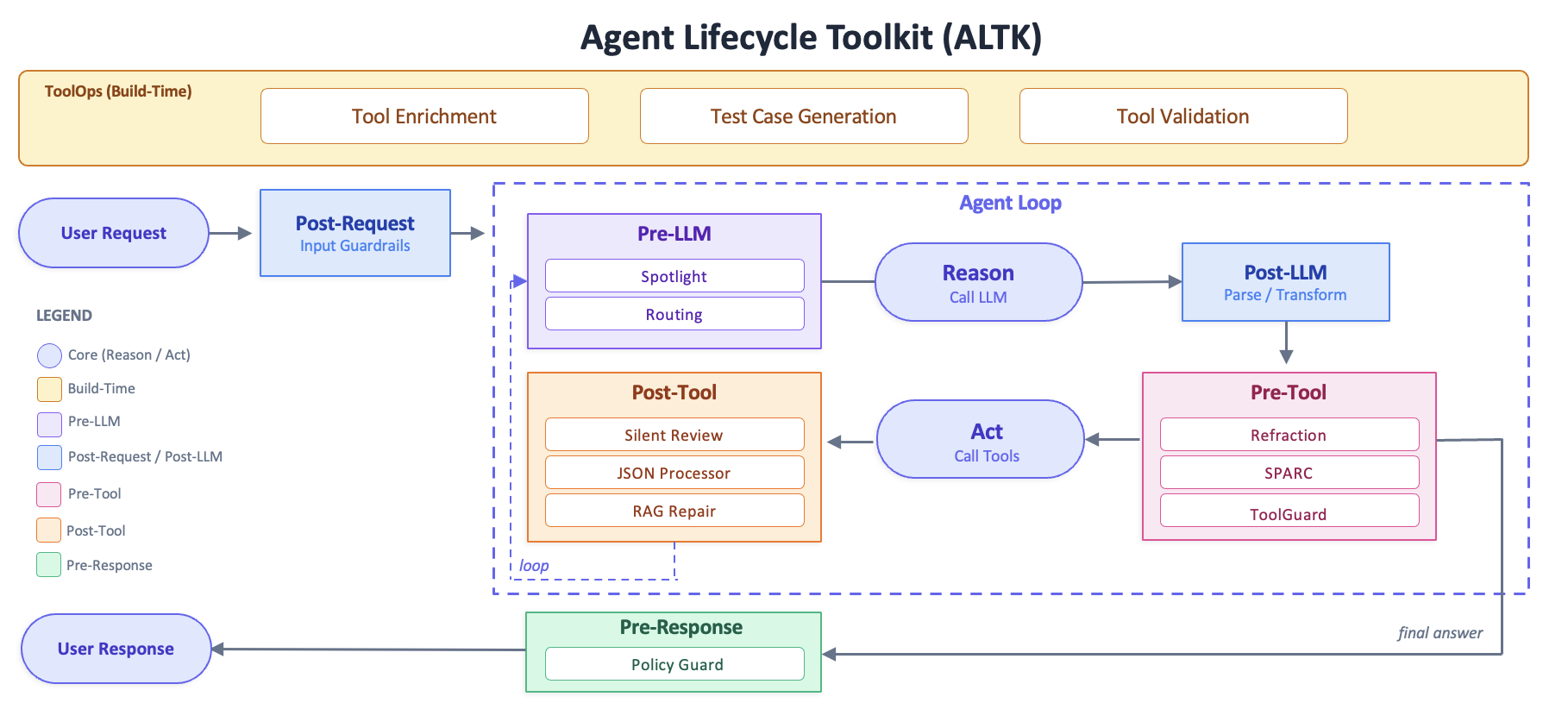}
    \caption{Agent lifecycle and corresponding ALTK components}
    \label{fig:lifecycle}
\end{figure*}

ALTK is organized around key stages in the agent lifecycle, as shown in Figure~\ref{fig:lifecycle}: build-time, pre-LLM prompt preparation, post-LLM, pre-tool call, post-tool call, and pre-final response. The main design goals are \emph{separation of concerns} and \emph{modularity}: each component targets one dominant error mode and can be enabled independently or in combination with others.

ALTK is currently implemented as an open source Python library called \emph{altk-boost} that provides a simple interface into these lifecycle components. We intend for this library and its components to be plug-and-play and as framework-agnostic as possible. Integrating an ALTK component into an agent during runtime has three phases which can be done in three lines of code: 1) defining the input to the component, 2) instantiating and configuring the component, and 3) processing the given input and reviewing the result. Concretely, for the code example in Figure~\ref{fig:code-example}, we use the Silent Error Review component in the ALTK as a post-tool hook to check for silent errors in a problematic tool. This component expects the current log of messages and tool response as input and then after instantiating the Silent Error Review component, these messages are processed and a result is given whether or not a silent error was detected. This result can be given back to the agent to prevent unintended behavior. Each component in the ALTK library follows the same basic interface and phases. The library is available on GitHub\footnote{https://github.com/AgentToolkit/altk-boost} and PyPi\footnote{https://pypi.org/project/altk-boost/}. For more code examples, please see the \emph{examples} folder in repository. A video walkthrough\footnote{https://www.youtube.com/watch?v=FsTuf9fmgM4} is available with additional demonstration videos on the ALTK YouTube channel\footnote{https://www.youtube.com/@AgentToolkit}.

\begin{figure}[t]
\centering
\small
\begin{minted}[breaklines,frame=lines,fontsize=\footnotesize]{python}
def post_tool_hook(state: AgentState) -> dict:
    # Post-tool node to check the output of a problematic tool
    tool_response = state["messages"][-1].tool_outputs
    review_input = SilentReviewRunInput(messages=state["messages"], tool_response=tool_response)
    reviewer = SilentReviewForJSONDataComponent()
    review_result = reviewer.process(data=review_input, phase=AgentPhase.RUNTIME)
    if review_result.outcome == Outcome.NOT_ACCOMPLISHED:
        return "Silent error detected, retry the tool!"
    else:
        # (allow tool call to go through)

agent_graph.add_edge("flaky_tool", "post_tool_hook")
\end{minted}
\caption{Code example in Python integrating the Silent Error Review component from ALTK to a LangGraph agent as a post-tool hook.}
\Description{Code example for integrating the Silent Error Review component from ALTK to a LangGraph agent as a post-tool hook.}
\label{fig:code-example}
\end{figure}

\begin{figure}
    \centering
    \includegraphics[width=\linewidth]{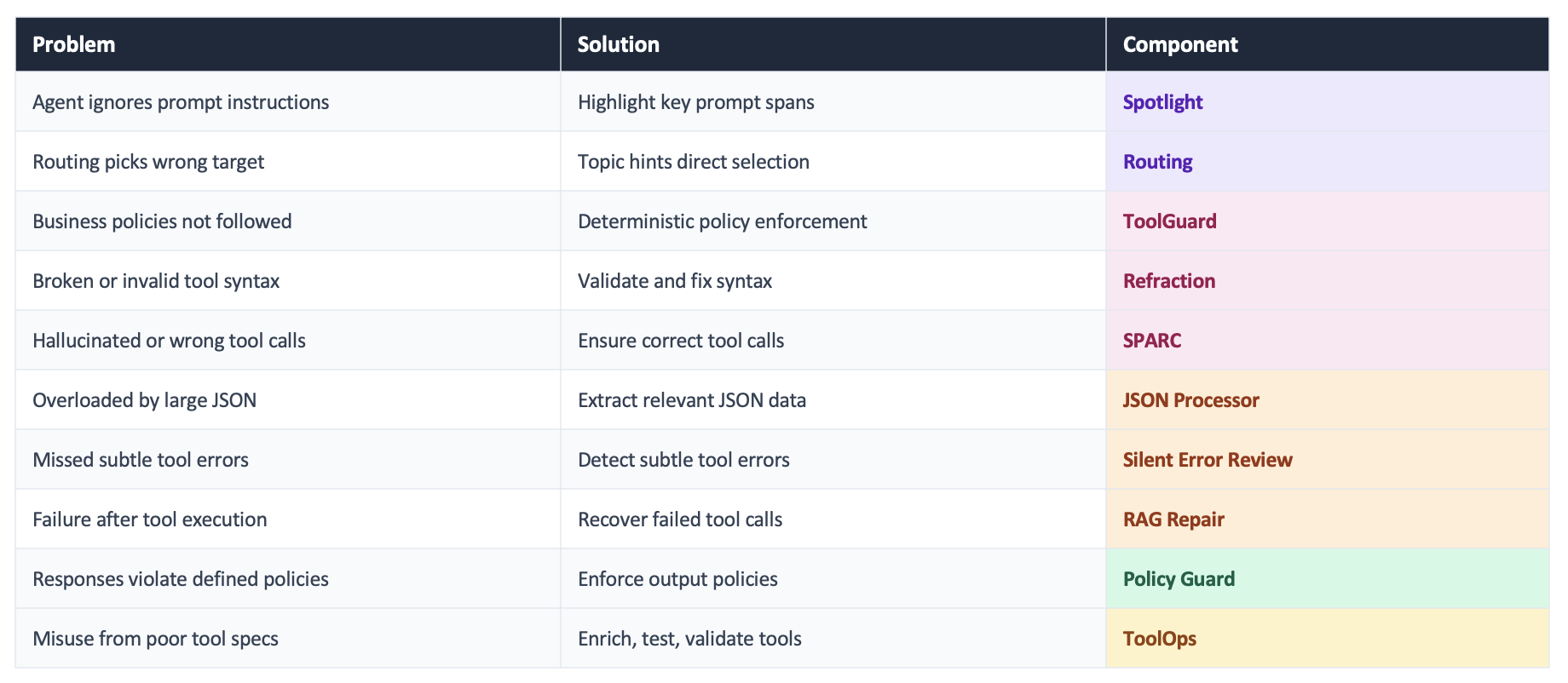}
    \caption{List of components in ALTK}
    \label{fig:components}
\end{figure}

At the time of writing, ALTK has 10 components (see Figure~\ref{fig:components}), spanning lifecycle stages from build-time to right before the response is given to the user. For this demonstration, we focus on three components: SPARC as a \emph{pre-tool} call validator, JSON Processor as a \emph{post-tool} processor of long responses, and a \emph{post-tool} Silent Error Review. 

\subsection{Pre-tool: SPARC - pre-execution validation}
In enterprise settings, semantically incorrect tool calls may waste API quota, corrupt external state, or trigger irreversible actions. Production agents need an inline runtime mechanism that decides whether a specific call should even be allowed to execute. 
SPARC is a component that works at the pre-tool lifecycle stage. Based on the message history, the provided tool specifications and any candidate tool calls, SPARC performs \textit{syntactic} validation, \textit{semantic} validation, and \textit{transformation} validation.
Syntactic validation is rule-based and catches non-existent tools, unknown arguments, missing required parameters, type mismatches, and JSON-schema violations. Semantic validation uses one or more LLM judges to assess function-selection appropriateness, parameter grounding, hallucinated values, value-format alignment, and unmet prerequisites. Transformation validation handles format or unit mismatches (for example, date, currency formats) and performs automatic conversions as demanded by the tool specifications.
The output of the component identifies whether tool call is invalid. If the tool call is not valid, SPARC identifies issues and suggests remediation. 

\subsection{Post-tool: JSON Processor}
In a tool-augmented agentic pipeline, the JSON processor acts as a critical middleware layer between raw API output and downstream reasoning. Passing voluminous, deeply nested JSON directly into the agent's context competes for attention with the task prompt and has been shown to degrade accuracy as responses grow longer~\cite{jsonprocessor}. Instead, this component delegates the parsing to a code generation step: the LLM is prompted to write a short Python function that navigates the JSON structure, applies any necessary filtering or aggregation logic, and returns only the extracted answer. This approach treats the LLM as a programmer rather than a reader, playing to its strength in producing structured code. When augmented with the API's JSON response schema, the generated parser becomes even more reliable; the model can reason about field names, data types, and nesting relationships without needing to infer them from the raw data. The resulting agent architecture is both more token efficient, since the code output is far smaller than the full response it processes, and more composable, as the deterministic output of an executed script slots cleanly into the next stage of an agent loop without any formatting noise and verbosity.

\subsection{Post-tool: Silent Error Review}
A common scenario for agentic tool usage is "soft failures" where an API may return responses that seem correct by returning a HTTP status code of "200 OK" but the body of the response may contain text like "Service under maintenance" or "No results found." Traditional agents often interpret this tool response as a correct final answer and this may cause unintended behavior. The Silent Error Review component works at the post-tool stage to identify these failures using a prompt-based approach. The component takes as input the user query, the tool response, and optionally the tool specification, to review the response as ``ACCOMPLISHED'', ``PARTIALLY ACCOMPLISHED'', or ``NOT ACCOMPLISHED''. 

\section{Evaluations}
\label{sec:evaluation}
As ALTK is comprised of many components, each component is individually evaluated for effectiveness in their particular task. We present evaluations for the subset of ALTK components we focus on for this demonstration. 

\subsection{Pre-tool: SPARC Evaluation}

\begin{figure}
    \centering
    \includegraphics[width=\linewidth]{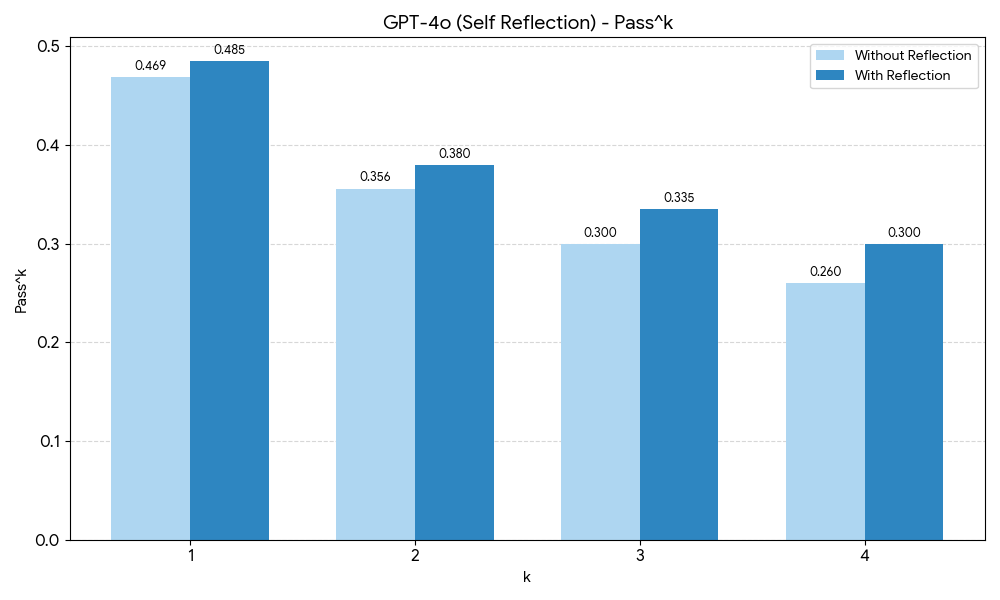}
    \caption{$\tau$-bench airline pass$^k$ with and without SPARC. ``With reflection'' inserts SPARC before each tool call and returns critique and/or corrections to the agent when a call is rejected.}
    \label{fig:taubench}
\end{figure}

We evaluate SPARC on the airline API subset of the $\tau$-bench dataset \cite{yao2024taubench} in a ReAct loop. If the candidate call is approved, it executes. If rejected, the reflection artifact (issue type, evidence, and correction suggestion) is fed back to the agent, which retries. 
Figure~\ref{fig:taubench} reports the experimental results. The main pattern is that gains grow with $k$. For GPT-4o self-reflection, pass$^1$ moves from 0.470 to 0.485, while pass$^4$ improves from 0.260 to 0.300. 
SPARC is most helpful for near-miss trajectories: it turns many incorrect first proposals into recoverable tool decisions on subsequent retries.

\subsection{Post-tool: JSON Processor}
\begin{figure}
    \centering
    \includegraphics[width=\linewidth]{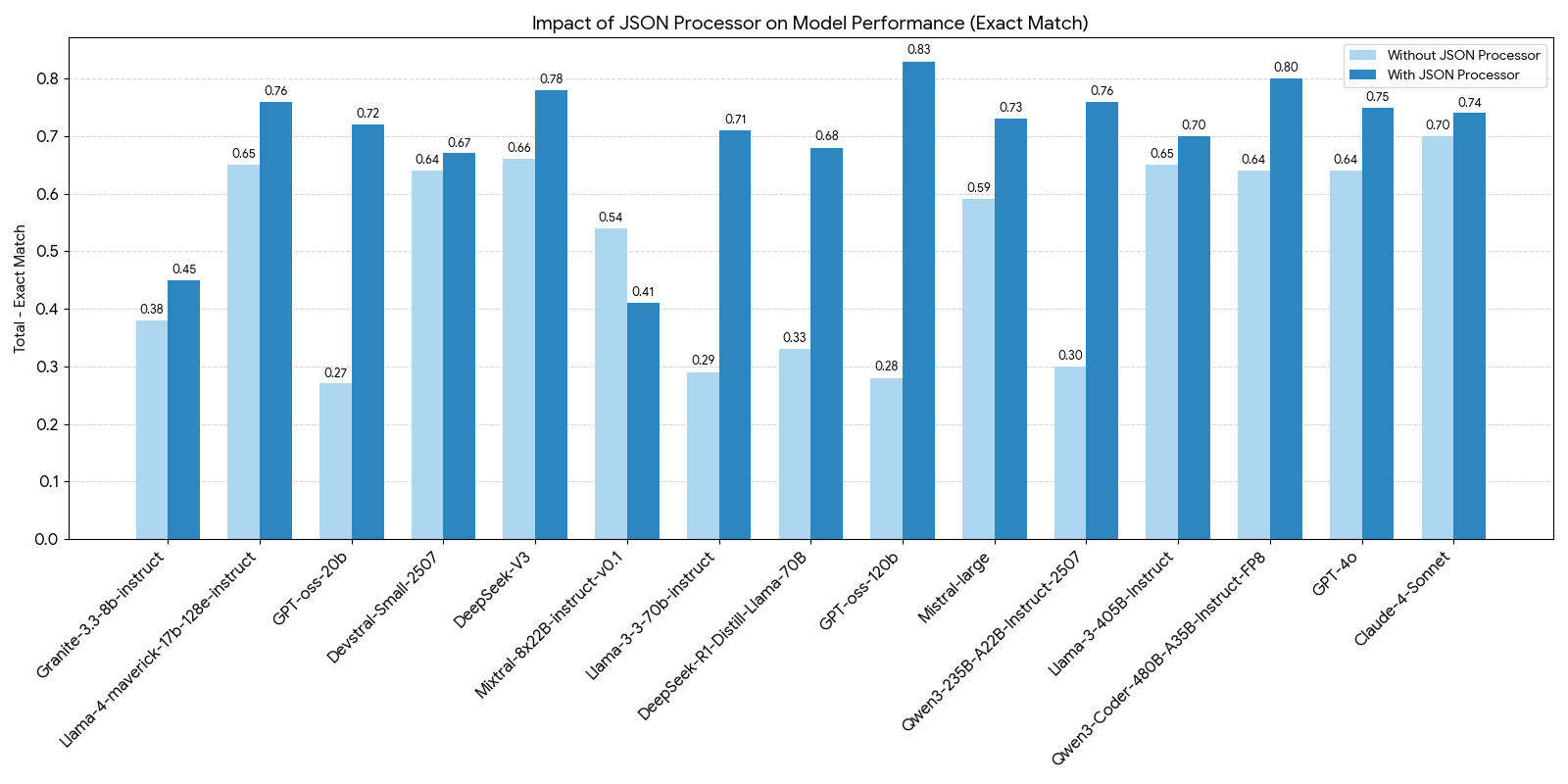}
    \caption{Model performance with and w/o JSON Processor\cite{jsonprocessor}}
    \label{fig:json_proc_res}
\end{figure}

We evaluate the JSON processor component on 15 models from various families and sizes on a dataset of approximately 1,300 JSON responses queries of varying complexity. Using the JSON processor leads to improvements over directly prompting a model to retrieve the answers without using the JSON processor component. 
Figure \ref{fig:json_proc_res} shows 16\% improvement on average across models when using the JSON processor \cite{jsonprocessor}.  

\subsection{Post-tool: Silent Error Review Evaluation}

\begin{figure}
    \centering
    \includegraphics[width=\linewidth]{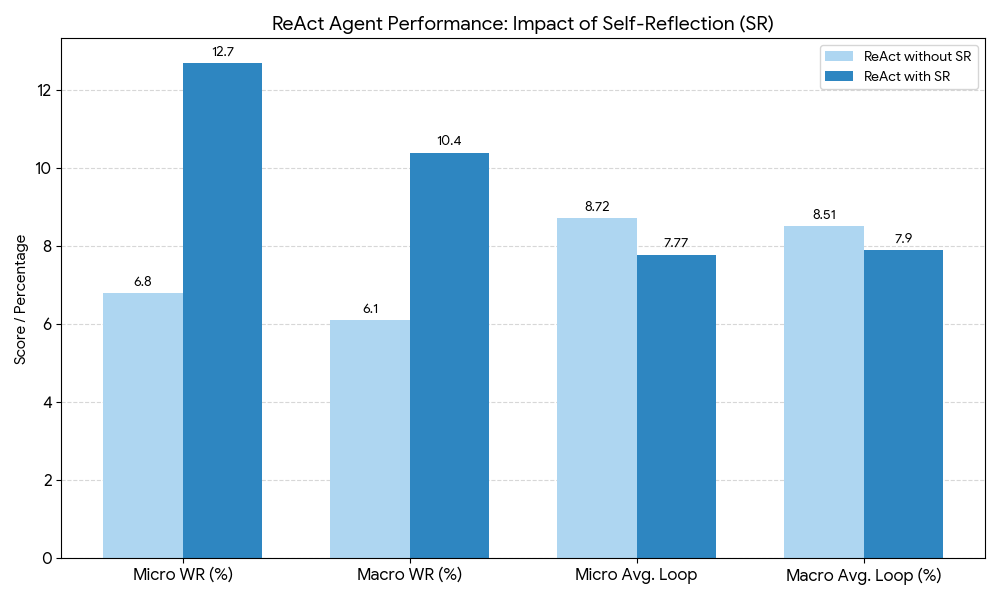}
    \caption{ReAct agent performance comparison with and without Silent Error Review on LiveAPIBench \cite{elder2026liveapibench2500live}; WR: Win Rate}
    \label{tab:fcrewardbench}
\end{figure}

We evaluate the Silent Error Review component on the LiveAPIBench dataset for SQL queries in a ReAct loop \cite{elder2026liveapibench2500live}. We evaluate using two metrics: Micro Win Rate and Macro Win Rate. The Micro Win Rate is the average performance across individual data subsets and the Macro Win Rate is the overall performance across all samples in all subsets. As seen in figure ~\ref{tab:fcrewardbench}, adding Silent Error Review to the ReAct loop nearly doubles the micro win rate, indicating more queries are fully or partially accomplished. Additionally, the average loop counts decrease showing that fewer iterations are needed to reach success.

\section{Usage and Integrations}
To maximize accessibility and facilitate adoption across diverse enterprise environments, ALTK is architected to support multiple deployment approaches. Rather than forcing a single integration pattern, the toolkit targets three distinct developer profiles: pro-code, low-code, and no-code. 


In the pro-code case, for software engineers and AI researchers requiring fine-grained control over the execution flow, ALTK is distributed as a modular set of open-source Python libraries hosted on GitHub.
Developers install the core packages and directly modify their existing agent source code, embedding ALTK’s lifecycle hooks (Pre-Tool, Post-Tool, Pre-Response) natively within their logic. There are several Jupyter notebooks with examples of how to use the various components\footnote{https://agenttoolkit.github.io/altk-boost/examples/}.



For rapid prototyping and visual workflow builders, ALTK provides deep integration with the Langflow\footnote{https://www.langflow.org/blog/langflow-1-7} low-code tool. 
ALTK capabilities are exposed via a custom agent within Langflow's visual authoring environment. The ToolGuard
, JSON Processor
, and SPARC
components are available in Langflow with demos on our YouTube channel\footnote{https://www.youtube.com/@AgentToolkit\label{fn:youtube}}.



For system administrators, platform engineers, and operations teams who need to secure and enhance legacy or third-party agents without altering their underlying code (no-code), ALTK integrates seamlessly as a middleware layer via the ContextForge MCP Gateway\footnote{https://ibm.github.io/mcp-context-forge/}.
Administrators configure the ALTK integrations in the gateway to intercept outbound tool calls and inbound responses, applying ALTK’s SPARC
or JSON Processor
dynamically. Tool enrichment and testing are available at the gateway for more effective agent tool use.
See our YouTube channel\footref{fn:youtube} for demos of these integrations. 


\section{Related Work}
A wide ecosystem of agent orchestration frameworks, such as LangChain \cite{chase2022langchain}, LangGraph \cite{langchain2024langgraph}, LlamaStack \cite{llama-stack}, LlamaIndex~\cite{liu2022llamaindex}, CrewAI~\cite{moura2023crewai}, AutoGen~\cite{wu2023autogen}, Claude’s Agent SDK \cite{anthropic2024sdk}, Bee~\cite{ibm2024bee}, Smolagents~\cite{huggingface2024smolagents}, and Haystack~\cite{moller2020haystack}, provides tooling to build LLM‑based agents. These frameworks largely focus on workflow composition, infrastructure, or developer‑side robustness. They offer hooks and middleware \cite{langchain2024middleware} for retries, fallbacks, and tool routing, but they generally do not detect or prevent semantic errors in agent reasoning or tool use.

ALTK fills this gap by acting as a framework‑agnostic reliability layer that integrates at lifecycle hook points within any agent system. 
For example, ALTK can plug into LangChain middleware or be exposed as skills or MCP components within Claude’s SDK.

Compared to data and model‑centric approaches (e.g., APIGen \cite{liu2024apigen}, ToolACE~\cite{liu2024toolace}, Granite function‑calling models~\cite{abdelaziz2024granitefc}), which improve average tool‑calling quality through better training, ALTK provides an inference‑time gate to determine whether a specific call should run.

Compared to reflection‑ and repair‑based methods (e.g., Reflexion~\cite{shinn2023reflexion}, REBACT~\cite{zeng2025rebact}, ToolReflection~\cite{polyakov2025toolreflection}, Failure Makes the Agent Stronger~\cite{su2025failure}, and Tool-MVR~\cite{ma2025toolmvr}), ALTK’s SPARC module is similar in spirit but differs by operating before execution, producing structured outputs rather than free‑form critiques, and combining semantic reflection with deterministic schema and execution‑ verification checks.

Overall, ALTK complements rather than replaces existing agent frameworks, providing systematic runtime safeguards that enhance correctness and reliability.

\section{Conclusion}
ALTK is motivated by a simple observation: robust agents need more than a capable base LLM; they need to address the dominant failure modes that occur at various points in the agent lifecycle. ALTK provides a flexible toolkit of components to address common problems at these lifecycle stages. This flexibility is open-ended, as we invite agent builders to extend the toolkit with their own solutions to problems as well as integrate components into agentic systems. We believe lifecycle-based components are key to building agents that are intelligent, reliable, and adaptable.

Beyond runtime reliability, ALTK can also support analytics and evaluation workflows by applying its lifecycle checks to agent trajectories, enabling fine‑grained analysis of failure modes and model behavior. These same components can provide structured signals for reward model training or tuning, turning ALTK’s reflectors into supervision signals that improve tool‑use fidelity and policy adherence.


\bibliographystyle{ACM-Reference-Format}

\bibliography{main}




\end{document}